\def\year{2020}\relax
\documentclass[letterpaper]{article} 
\usepackage{aaai20}  
\usepackage{times}  
\usepackage{helvet} 
\usepackage{courier}  
\usepackage[hyphens]{url}  
\usepackage{graphicx} 
\usepackage{graphicx}  
\usepackage{amsmath}
\newcommand{\dogimg}[1]{\includegraphics[width=0.088\linewidth]{#1}}

\usepackage[noend]{algpseudocode}
\usepackage{here}
\usepackage{multirow}
\usepackage{comment}
\usepackage{array}
\usepackage{booktabs}
\usepackage{bm}
\usepackage{bbding}
\usepackage{pifont}
\usepackage{wasysym}
\usepackage{amssymb}
\usepackage{color}

\title{Block-wise Scrambled Image Recognition Using Adaptation Network}

\author{
Koki Madono\textsuperscript{\rm 1,2},
Masayuki Tanaka\textsuperscript{\rm 2},
Masaki Onishi\textsuperscript{\rm 2},
Tetsuji Ogawa\textsuperscript{\rm 1,2} 
\\ 
\textsuperscript{\rm 1}Department of Communications and Computer Engineering, Waseda University\\
\textsuperscript{\rm 2}The National Institute of Advanced Industrial Science and Technology\\
}

\begin{document}

\maketitle
\begin{abstract}
In this study,
a perceptually hidden object-recognition method
is investigated
to generate secure images 
recognizable by humans but not machines.
Hence, 
both the perceptual information hiding and the corresponding object recognition methods should be developed.
Block-wise image scrambling is introduced
to hide perceptual information 
from a third party.
In addition,
an adaptation network is proposed
to recognize those scrambled images.
Experimental comparisons conducted using CIFAR datasets demonstrated that
the proposed adaptation network performed well
in incorporating simple perceptual information hiding into DNN-based image classification.
\end{abstract}

\section{Introduction}
\label{sec:intro}

Cloud-based image analysis services,
such as Google Cloud~\cite{gcloud}
and Microsoft Azure~\cite{azure},
have become extremely powerful and easy to use.
These systems, however, can be further improved
in terms of privacy issues.
For example,
when a client transfers an image to a cloud server,
a third party can view the image.
It is noteworthy that
encrypted communications or secure communications are insufficient
because the image data should be decrypted
to analyze at the cloud.
Deep neural networks (DNNs) have been widely used
in image processing~\cite{LeCun2015DeepL,Lowe1999ObjectRF,He2015DeepRL}; they are important in cloud-based image analysis.
In general, DNNs require a large number of images for training.
However, it becomes a critical problem
when the cloud-based image analysis is conducted 
based on privacy-sensitive images,
because the client has to provide such images
to a third party for machine learning.
Homomorphic encryption~\cite{Lagendijk2013EncryptedSP,Lu2016UsingFH} may address such a problem.
However, it is not feasible 
from the viewpoints of computational and memory costs.
Furthermore, mathematical operations in homomorphic encryptions are limited.

Table~\ref{fig:accessibility_relation} summarizes the accessibility of humans and machines to perceptual information.
Both humans and machines can access plain images.
Although these images are easy to use, 
attackers can easily access them.
Meanwhile,
homographic encryptions can hide perceptual information;
however, in general, it is difficult to use 
as aforementioned.
Therefore, 
in the present study,
a framework
for hiding perceptual information
from humans 
while machines continue accessing the information is developed.
\begin{table}[t]
    \caption{Accessibility of human and machine to perceptual information. Difficulty of attacks is also listed as well.}
    \label{fig:accessibility_relation}
    \vspace{1mm}
    \centering
    \begin{tabular}{c|ccc}
    \toprule
        &
        plain &
        scrambled &
        homographic \\
       &
      image &
      image &
      encryption \\
     \midrule
    human    &  $\checkmark$ &  & \\       
    machine  &  $\checkmark$ & $\checkmark$ & \\ 
    \midrule
    attacker &  easy & medium & difficult\\ 
    \bottomrule
     \end{tabular}
     \vspace{-10mm}
\end{table}
\begin{figure*}[t]
    \begin{center}
            \vspace{-12mm}
            \includegraphics[width=\linewidth]{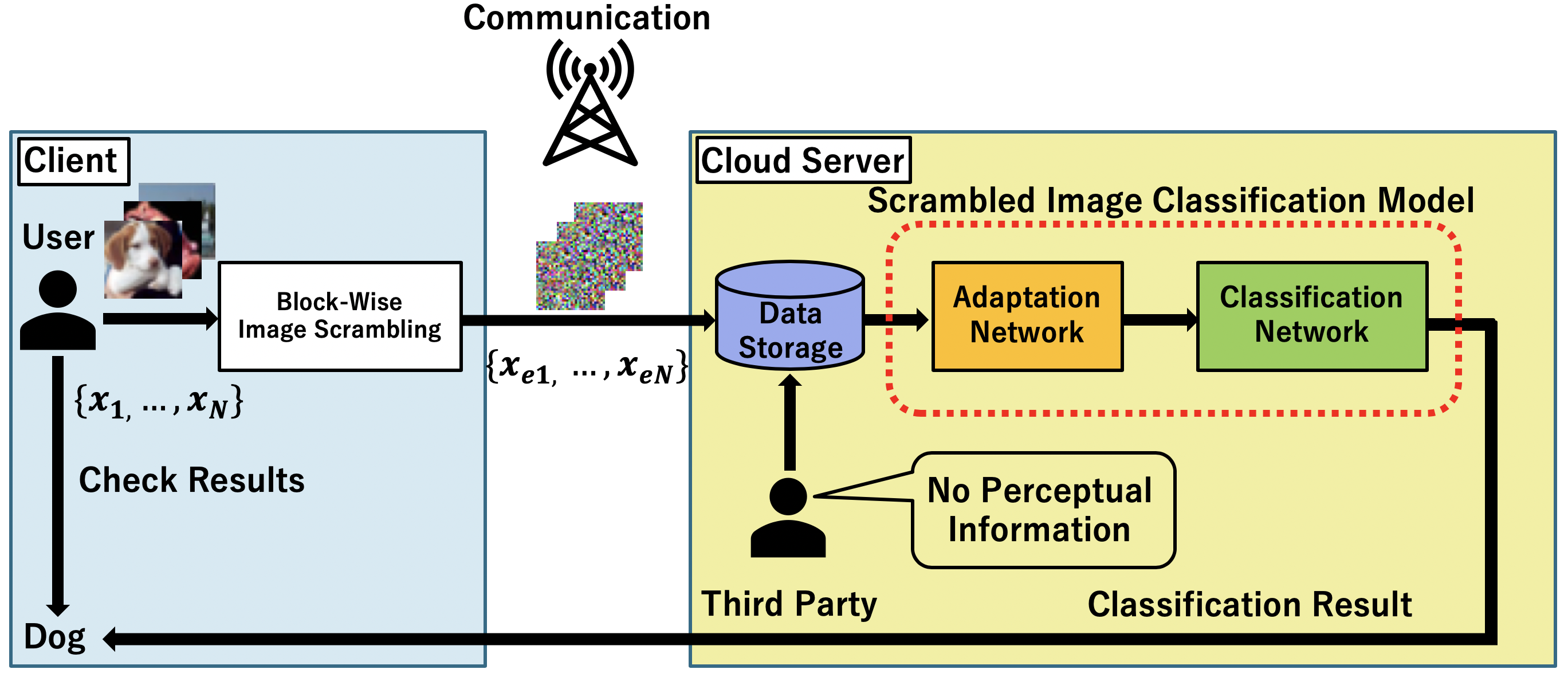}
            \vspace{-9mm}
            \caption{Conceptual image of cloud-based image analysis with image scrambling. Client can access the original image and object class in it. Third party for image analysis can access object class and perceptually hidden images for training image classifiers on cloud.}
            \label{fig:cloud_dnn}
            \vspace{-3mm}
    \end{center}
\end{figure*}
\begin{figure*}[t]
  \begin{center}
            \includegraphics[width=\linewidth]{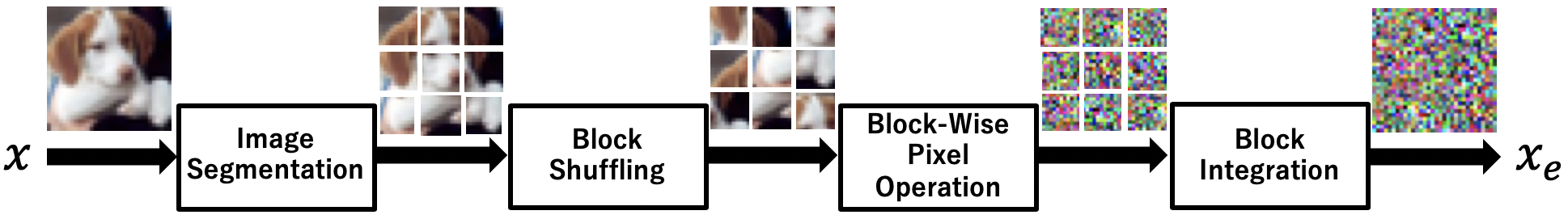}
            \vspace{-8mm}
            \caption{Processing pipeline of block-wise image scrambling. Image is segmented into blocks with $B \!\times\! B$ pixels. Block locations are shuffled, pixels in block are shuffled independently in every block, and shuffled blocks are concatenated.}
            \label{fig:encryption_scheme}
            \vspace{-8mm}
  \end{center}
\end{figure*}

Hence,
the present study focuses on block-wise image scrambling
as a simple perceptual information hiding method.
Several studies regarding block-wise image scrambling have been conducted,
such as learnable encryption (LE)~\cite{Tanaka2018LearnableIE}
and encryption-then-compression (EtC)~\cite{Chuman2018EncryptionThenCompressionSU}.
It is noteworthy that 
the scrambled images are not exactly encrypted
although the perceptual information can be hidden.
In addition,
existing DNN architectures,
which are proven effective 
in image recognition~\cite{He2015DeepRL,Simonyan2014VeryDC,Yamada2018ShakeDropRF},
do not assume the scrambled images as inputs,
indicating that another component is required
to recognize such images.

Hence,
in this study, the LE algorithm
is extended to increase the security level of perceptually hidden images
and introduce the adaptation network
for recognizing the scrambled images.
Experimental verification conducted using CIFAR datasets~\cite{Lowe1999ObjectRF} demonstrates that
images with block-wise scrambling can be recognized
by integrating the proposed adaptation network
with the existing DNN-based classifier~\cite{He2015DeepRL,Simonyan2014VeryDC,Yamada2018ShakeDropRF}.
Insights from the present study can contribute to the generation of images
recognizable by humans but not machines.


\section{Perceptual Information Hiding}
\label{sec:protection}
Perceptual information hiding,
in which perceptual information is hidden from humans
but can be accessed by machines,
is introduced
to DNN-based image recognition.
Figure~\ref{fig:cloud_dnn} illustrates an assumed cloud-based image recognition,
in which perceptually hidden images are obtained
by block-wise image scrambling
and used as inputs to a classifier 
on a cloud.
During the classifier training,
a client uploads scrambled images 
together with annotations.
A third party of model developer then trains the model
using the uploaded images.
Because naive DNN-based image classifiers do not use scrambled images as inputs,
the adaptation network is introduced
to manage such concealed data.

Once the scrambled image classification models are trained,
the assumed service can be used
while hiding the perceptual information
from a third party.
The third party, however, can analyze the adaptation network
to understand the scrambling process.
Therefore,
the use of adaptation networks is not perfect
in perceptual information hiding
but still offers some advantages
because effective algorithms for reconstructing original images 
from block-wise scrambled images are still available, 
to the best of our knowledge.


\section{Block-Wise Image Scrambling}
\label{sec:scrambling}

\begin{table*}[t]
  \vspace{-3mm}
  \caption{Summary of block-wise scrambling algorithms.
  Here, each block has $B \!\times\! B$ pixels and the number of blocks is $N$.}
  \label{tbl:dif-enc-scheme}
  \vspace{1mm}
  \centering
  \begin{tabular}{c|cccc}
    \toprule
      & plain image & LE~\cite{Tanaka2018LearnableIE} & EtC~(Chuman {\it et al.} 2018) & ELE\\
       &  &  &  & (proposed)\\
      \midrule
    image example & 
    \dogimg{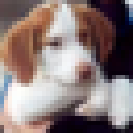} & \dogimg{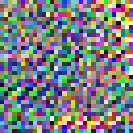} &
    \dogimg{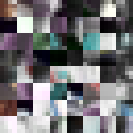} & \dogimg{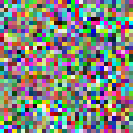} \\
    \hline
     &  &  &  & \\
    block key &  & common & different & different \\
     &  &  &  & \\
    \hline
     &  &  &  & \\
    \footnotesize{block shuffling} &  &  & $\checkmark$ & $\checkmark$ \\
     &  &  &  & \\
    \hline
     & & pixel shuffling, & {\footnotesize block rotation \& inversion,} & \footnotesize{pixel shuffling}, \\
    block-wise pixel operation & & {\footnotesize negative-positive transform} &
    {\footnotesize negative-positive transform,} & {\footnotesize negative-positive transform}\\
      & &  & {\footnotesize color component shuffling} & \\
      \hline
      &  & & & \\
      key space & $0$ &
      $(B^2 \cdot 6)! \cdot 2^{B^2 \cdot 6}$ &
      $8^{N} \cdot 2^{N} \cdot 6^{N} \cdot N!$ &
      \{$(B^2 \cdot 6)! \cdot 2^{B^2 \cdot 6}\}^{N} \cdot N!$\\
    & & & & \\
        \bottomrule
  \end{tabular}
\end{table*}

This section describes the processing pipeline
of block-wise image scrambling~\cite{Tanaka2018LearnableIE,Chuman2018EncryptionThenCompressionSU}.
Figure~\ref{fig:encryption_scheme} illustrates the pipeline.
In the developed system,
an input image is first divided into blocks
with $B \!\times\! B$ pixels,
where the number of yielded blocks is $N$.
For example,
the segmentation of a $32 \!\times\! 32$ pixel image into blocks with $4 \!\times\! 4$ pixels yields 64 blocks.
The positions of the segmented blocks are then shuffled (i.e., block shuffling).
In addition,
pixels in each block are shuffled
with security keys,
where different keys are applied
to every blocks
(i.e., block-wise pixel shuffling).
Finally,
the blocks with the shuffled pixels are concatenated
to obtain the resulting scrambled image.

Table~\ref{tbl:dif-enc-scheme} summarizes 
block-wise image scrambling algorithms.
The developed block-wise scrambling 
can be regarded
as an extension of the existing 
LE algorithm~\cite{Tanaka2018LearnableIE},
Therefore, the developed algorithm is referred 
to as the extended learnable encryption (ELE).


\subsection{Security Evaluation}
\label{ssec:securityeval}

This subsection discusses the security levels 
of the block-wise image scrambling algorithms
using the key space,
which refers to the set of all possible permutations of a key.

The LE shuffles pixels 
and negative--positive transforms
with the same key in every block.
Each block is split
into four upper bits and four lower bits 
to generate three to six channels of blocks.
Subsequently,
the pixels are randomly shuffled,
yielding $N_{\rm ps}$ combinations.
The intensity of randomly selected 
pixels are then reversed,
yielding $N_{\rm np}$ combinations.
The key space of the LE is calculated as
\begin{eqnarray}
    O({\rm LE}) &=& N_{\rm ps} \cdot N_{\rm np} \nonumber \\
    &=& (4^{2} \cdot 6)! \cdot 2^{4^{2} \cdot 6} \nonumber \\
    &=& 96! \cdot 2^{96}.
\end{eqnarray}

The EtC comprises 
block rotation followed by inversion ($N_{\rm r\&i}$), 
negative--positive transform ($N_{\rm np}$), 
color component shuffling ($N_{\rm col}$), 
and block location shuffling ($N_{\rm bs}$)
with the same key in every block.
The key space of the EtC is calculated as
\begin{eqnarray}
    O({\rm EtC}) &=& N_{\rm r\&i} \cdot N_{\rm np} \cdot N_{\rm col} \cdot N_{\rm bs} \nonumber \\
           &=& (4\cdot2)^{64} \cdot 2^{64} \cdot (3!)^{64} \cdot 64! \nonumber \\
           &=& 8^{64} \cdot 2^{64} \cdot 6^{64} \cdot 64!.
\end{eqnarray}

The ELE comprises block-wise pixel shuffling 
with different keys in every block ($N_{\rm dps}$) and block location shuffling ($N_{\rm bs}$).
The key space of the ELE is calculated as
\begin{eqnarray}
    O({\rm ELE}) &=& N_{\rm dps} \cdot N_{\rm bs} \nonumber \\
    &=&\{(4^{2} \cdot 6)! \cdot 2^{4^{2} \cdot 6}\}^{64} \cdot 64! \nonumber \\
    &=&(96! \cdot 2^{96})^{64}  \cdot 64!
\end{eqnarray}
Eventually,
the key spaces for the LE, EtC, and ELE are as follows:
\begin{equation}                    
    O({\rm ELE}) \gg O({\rm EtC}) \gg O({\rm LE}).
\end{equation}
This indicates that the ELE has a larger key space
than the LE and EtC.
\begin{figure*}[h]
\begin{center}
\includegraphics[width=\linewidth]{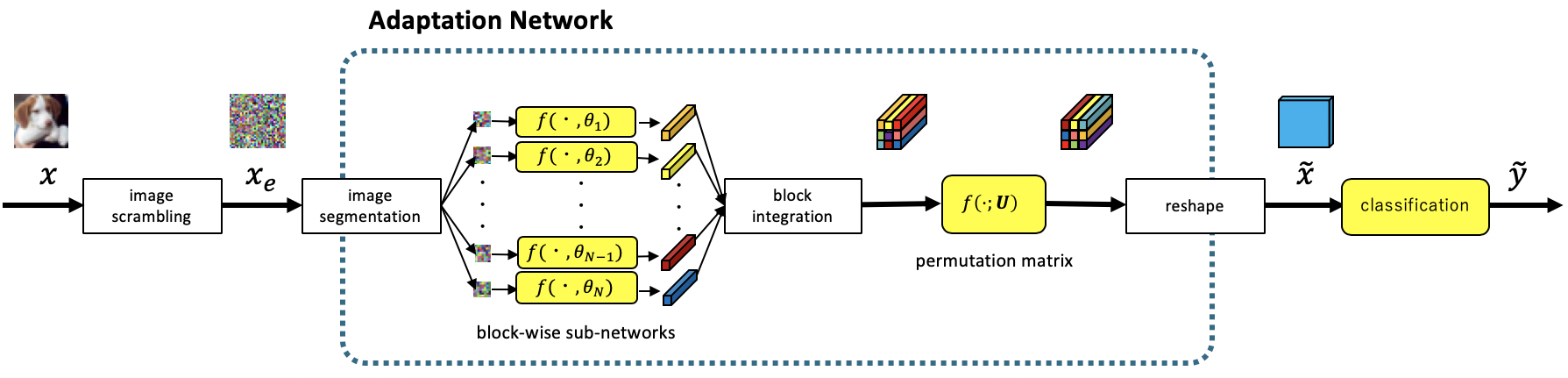}
\vspace{-7mm}
\caption{Network architecture of proposed adaptation network. Scrambled image is segmented
into blocks with $B \!\times\! B$ pixels;
feature maps of block-wise sub-networks, $\{f(\bm{x}_{e_1};\bm{\theta}_{1}),\cdots,f(\bm{x}_{e_N};\bm{\theta}_{N})\}$,
are integreted;
pseudo permutation matrix is applied
to integrated feature map; and
feature map is upscaled
by pixel shuffle layer~\cite{pixelshuffle}
and then used as input to classification network
based on shakedrop.
Colored boxes mean trainable processing.}
\vspace{-5mm}
\label{fig:model_overview}
\end{center}
\end{figure*}

\begin{figure*}[t]
    \begin{center}
            \includegraphics[width=\linewidth]{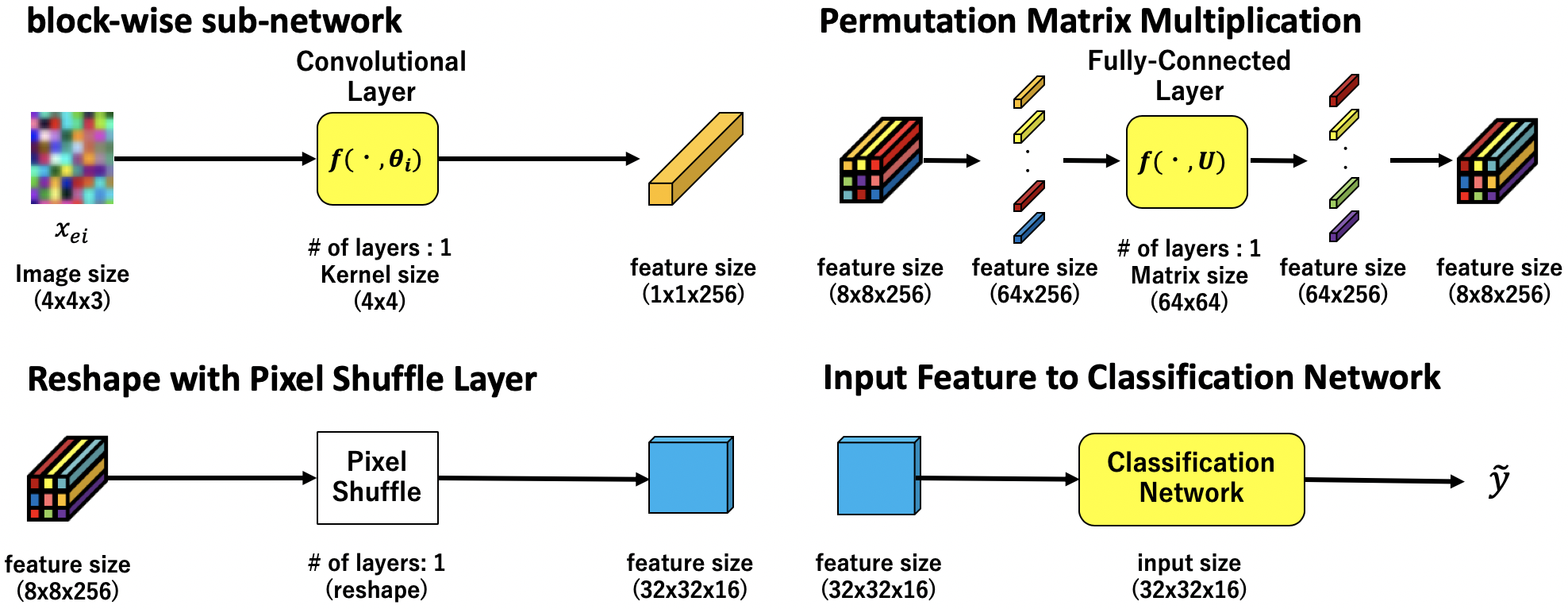}
             \vspace{-8mm}
            \caption{Detailed network architectures used. 
            Block size and kernel size are the same in every block-wise sub-networks. 
            Pseudo permutation matrix is applied
            to integrated feature map;
            Feature map is upscaled
            by pixel shuffle layer~\cite{pixelshuffle};
            and feature map is used as input to classification network~\cite{Yamada2018ShakeDropRF}.
            Colored boxes express trainable processing.}
            \label{fig:network_component}
            \vspace{-5mm}
    \end{center}
\end{figure*}

\section{Adaptation Network for Block-Wise Scrambled Image Recognition}
\label{sec:adaptnet}

In this section, proposes the adaptation network
to recognize the block-wise scrambled images,
considering the processing pipeline
of block-wise scrambling
as shown in Fig.~\ref{fig:encryption_scheme}.

The proposed adaptation network
comprises three parts:
block-wise sub-networks,
a learnable pseudo permutation matrix,
and a pixel shuffling layer~\cite{pixelshuffle}.
The schematic diagram of the developed system 
and detailed information 
on network architecture 
for each component are illustrated 
in Figs.~\ref{fig:model_overview} and \ref{fig:network_component},
respectively.

First,
a scrambled image $\bm{x}_e$ is segmented
into $N$ blocks with $B \!\times\! B$ pixels as
$\{\bm{x}_{e_1}, \bm{x}_{e_2}, \cdots, \bm{x}_{e_N}\}$,
where $\bm{x}_{e_b}$ represents a block (i.e., segmented image).
Each block is transformed 
by the corresponding block-wise sub-network $f(\bm{x}_{e_b};\bm{\theta}_b)$,
which is separately trained 
for each block.
This process aims at managing ELE-based image scrambling
i.e., block-wise scrambling with different keys.
The feature maps of the block-wise sub-networks are then integrated
and the pseudo permutation matrix is applied 
to the integrated feature map.
This operation corresponds to inverse block shuffling.
Subsequently,
the shuffled pixels are aligned 
on the pixel shuffling layer
and then used as an input to the classification network.
\begin{table*}[t]
\vspace{-5mm}
  \caption{Accuracy of scrambled image classification.}
  \label{tbl:result-classification}
  \vspace{1mm}
  \centering
  \begin{tabular}{ll|cccc}
    \toprule
      &  & plain image & LE~\cite{Tanaka2018LearnableIE} &  EtC~(Chuman {\it et al.} 2018) & ELE \\
     Dataset & Adaptation network &  &  &  & (proposed) \\     
      \midrule
    cifar-10 & No AdaptNet & {\bf 96.70\%} & {\bf 94.94\%} & 85.94\% & 67.10\% \\
    & LE-AdaptNet & 95.64\%  & 94.49\% &  80.16\% & 48.39\%\\
    & ELE-AdaptNet & 85.32\%  & 87.28\% & {\bf 89.09\%} & {\bf 83.06\%} \\ 
    \hline
    cifar-100 & No AdaptNet & {\bf 83.59\%} & {\bf 78.25\%} & 61.90\% & 43.05\%\\
    & LE-AdaptNet & 79.13\% & 75.48\% & 44.83\% &  7.19\%\\
    & ELE-AdaptNet & 60.36\% & 71.30\%& {\bf 71.91\%} &{\bf 62.97\%}\\
        \bottomrule
  \end{tabular}
\end{table*}
\begin{table*}[t]
  \vspace{-5mm}
  \caption{Estimates and their posterior probabilities of scrambled image classification using different models.}
  \label{tbl:result-class_probability_estimation}
  \vspace{1mm}
  \centering
  \begin{tabular}{l|cccc}
    \toprule
    & plain image & LE~\cite{Tanaka2018LearnableIE} &  EtC~(Chuman {\it et al.} 2018) & ELE\\
    Adaptation Network&  & &  & (proposed)\\
    \midrule
      & \dogimg{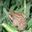} & \dogimg{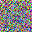} &  \dogimg{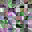} & \dogimg{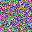} \\
      Shakedrop & frog(1.00) & frog(1.00) & frog(1.00) & deer(0.65) \\
      LE-AdaptNet + Shakedrop & frog(1.00) & frog(1.00) & frog(0.99) & deer(0.63) \\
      ELE-AdaptNet + Shakedrop  & frog(1.00) & frog(1.00) & frog(1.00) & frog(1.00) \\
        \midrule
      & \dogimg{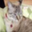} & \dogimg{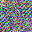} &  \dogimg{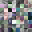} & \dogimg{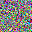} \\
      Shakedrop &  cat(1.00)& cat(1.00)  & cat(0.99) & bird(0.28) \\
      LE-AdaptNet + Shakedrop & cat(1.00) & cat(0.99) & cat(0.99) & bird(0.34) \\
      ELE-AdaptNet + Shakedrop  & cat(1.00) & cat(1.00) & cat(1.00) & cat(1.00) \\
      \midrule
      & \dogimg{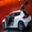} & \dogimg{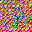} &  \dogimg{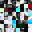} & \dogimg{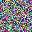} \\
      Shakedrop & automobile(1.00) & automobile(1.00) & automobile(0.98) & automobile(0.44) \\
      LE-AdaptNet + Shakedrop & automobile(1.00) & automobile(1.00) & automobile(0.99) & dog(0.55) \\
      ELE-AdaptNet + Shakedrop  & automobile(0.97) & automobile(0.94)& automobile(1.00) & automobile(0.82) \\
     \bottomrule
  \end{tabular}
\end{table*}
In theory,
if the permutation matrix for block shuffling is known,
its inverse matrix can be used to solve block shuffling.
Such a permutation matrix, however, is unknown.
Instead of estimating the inverse permutation matrix, 
we introduce pseudo permutation
and train the corresponding matrix ${\bm U}$ 
such that its elements can be sparse.
Hence,
the $L_{1-2}$ penalty~\cite{Lyu2019AutoShuffleNetLP,Esser2013AMF,Yin2015MinimizationO} is added to the loss function for training.
In addition,
all elements of $U$ should be non-negative
and the sum of each column and that of each row should be one.
Such constraints, however, are not considered herein.

The feature map of the plain image is expected to be spatially smooth. 
Additionally, the feature map of the adaptation network should exhibit the same characteristics
as those of the natural image.
Therefore,
the penalty with respect to spatial smoothness is added to the loss function.

The resulting loss function for training the adaptation network is written as
\begin{equation}
    L = L_{CE} + \lambda_U L_{U} + \lambda_s L_{s}\,,
\end{equation}
where $L_{CE}$ denotes the cross entorpy loss
for classification~\cite{Mannor2005TheCE},
$L_{U}$ denotes the $L_{1-2}$ penalty
for the pseudo permutation matrix,
and $L_{s}$ denotes the spatial smoothness penalty
for the feature map of the adaptation network.
Here,
$\lambda_U$ and $\lambda_s$ were empirically determined
as $\lambda_U=0.001$ and $\lambda_s=0.1$, respectively.

The cross entropy loss $L_{CE}$ is computed as
\begin{equation}
    L_{CE} =  - \frac{1}{M} \sum_{m=1}^M \sum_{k=1}^{K} t_{m,k} \log  C_k(\widetilde{\bm{x}}_m;\bm{\theta}_{c}) \,,
\end{equation}
where $m$ and $k$ denote the sample and class indexes,
respectively;
$t_{m,k}$ denotes the one-hot encoded label;
$C_k(\widetilde{\bm{x}}_m;\bm{\theta}_{c})$ denotes the posterior probability for the $k$-th class
of the classification network;
$\widetilde{\bm{x}}_m$ denotes the feature map 
of the adaptation network
for the $m$-th sample; and
$\bm{\theta}_{c}$ denotes the parameters 
of the classification network.

The $L_{1-2}$ penalty for the pseudo permutation matrix $L_{U}$ is calculated as
\begin{equation}
    \begin{split}
            L_{U} = \frac{1}{N \times N} \left\{ \sum_{i=1}^{N} \left[ \sum_{j=1}^{N}|u_{i,j}| -  \left(\sum_{j=1}^{N}u_{i,j}^{2} \right)^{1/2} \right] + \right. \\ 
            \left. \sum_{j=1}^{N} \left[ \sum_{i=1}^{N}|u_{i,j}| -  \left(\sum_{i=1}^{N}u_{i,j}^{2}\right)^{1/2} \right] \right\}
 \,,
    \end{split}
\end{equation}
where $u_{i,j}$ denotes the element of the matrix $U$
whose size is $N \!\times\! N$.

The spatial smoothness penalty of the feature map $L_{s}$ is calculated as
\begin{equation}
 \begin{split}
  L_s &= \frac{1}{M} \sum_{M=1}^{M} \left[ L_{s,i}^{(H)} + L_{s,i}^{(V)} \right] \\
  L_{s,i}^{(H)} &= \frac{1}{H(W-1)C} \sum_{h=1}^{H} \sum_{w=1}^{W-1} \sum_{c=1}^{C} 
[\widetilde{x}_i^{(H)}(h,w,c)]^2 \\
  L_{s,i}^{(V)} &= \frac{1}{(H-1)WC} \sum_{h=1}^{H-1} \sum_{w=1}^{W} \sum_{c=1}^{C} 
[\widetilde{x}_i^{(V)}(h,w,c)]^2 \,,\\
 \end{split}
\end{equation}
where $H$, $W$, and $C$ denote the height, width, and number of channels
in the feature map, respectively;
$\widetilde{\bm{x}}_i^{(H)}(h,w,c)$ and $\widetilde{\bm{x}}_i^{(V)}(h,w,c)$ denote the horizontal and vertical forward difference
of the $i$-th feature
at $(h,w,c)$, respectively.

\section{Experimental Validation}
\label{sec:exp}

Experimental validation was conducted
using CIFAR datasets
to demonstrate the effectiveness 
of the proposed scrambled image recognition method
\footnote{\color{magenta}https://github.com/MADONOKOUKI/Block-wise-Scrambled-Image-Recognition}.
Three systems using the following adaptation networks were compared as follows:
\begin{itemize}
\setlength{\parskip}{0cm}
\setlength{\itemsep}{0cm}
\item {\bf No AdaptNet}: No adaptation network
\item {\bf LE-AdaptNet}: Adaptation network with block independent (shared) sub-networks~\cite{Tanaka2018LearnableIE}
\item {\bf ELE-AdaptNet}: Adaptation network with block dependent sub-networks and block permutation matrix (proposed)
\end{itemize}
Because the existing {\bf LE-AdaptNet}~\cite{Tanaka2018LearnableIE} considers LE-based block-wise image scrambling,
it shares the block-wise network among all blocks
and does not require a pseudo permutation matrix.

The plain image and three types of scrambled images were evaluated as follows:
\begin{itemize}
\setlength{\parskip}{0cm}
\setlength{\itemsep}{0cm}
\item {\bf plain image} (i.e., no scrambling);
\item learnable encryption ({\bf LE})~\cite{Tanaka2018LearnableIE};
\item encryption-then-compression ({\bf EtC})~\cite{Chuman2018EncryptionThenCompressionSU}; and
\item extended learnable encryption ({\bf ELE}) 
proposed in this study.
\end{itemize}

The shakedrop network~\cite{Yamada2018ShakeDropRF} was exploited for the classification network
with the same setting as that of the original implementation.

\subsection{Experimental Setup}
\label{ssec:exp-setup}

The datasets used for the evaluation were CIFAR-10 and CIFAR-100~\cite{Lowe1999ObjectRF}.
All images were converted into block-wise scrambled images
and used as inputs
to the adaptation network.
The data augmentation was performed 
before block-wise scrambling.
The mini-batch size was 128 during training and testing.
The SGD with the Nesterov was used 
as the optimizer,
where the momentum was 0.9.
The network was trained with 300 epochs of iterations.
The learning rate was scheduled as 0.1 for 0-to-150 epochs,
0.01 for 150-to-225 epochs,
and 0.001 for 225-to-300 epochs.

\subsection{Experimental Results}
\label{ssec:exp-result}


Table~\ref{tbl:result-classification} lists the accuracy 
of recognizing the plain images and three types of scrambled images,
where the boldface represents the best accuracy.
The proposed {\bf ELE-AdaptNet} yielded the best accuracy 
for ELE- and EtC-based image scrambling.
For LE-based image scrambling,
the {\bf LE-AdaptNet}~\cite{Tanaka2018LearnableIE} performed better than the {\bf ELE-AdaptNet}
because the {\bf LE-AdaptNet} is specialized 
to LE-based image scrambling.
However, it is noteworthy that
the LE-based image scrambling is worse in terms of security
than the ELE-based image scrambling,
as shown in Table~\ref{tbl:dif-enc-scheme}.
Either the LE or ELE can be selected
as a perceptual information hiding method,
considering the tradeoff
between security level and recognition accuracy.
If the security level is prioritized,
the ELE-based approach should be selected.

Table~\ref{tbl:result-class_probability_estimation} lists examples of the plain and scrambled images
with the best matching object classes and corresponding posterior probabilities 
in parentheses.
For the plain image and LE- and EtC-based scrambled images,
the posterior probabilities are confident,
regardless of the adaptation networks.
For ELE-based image scrambling,
the proposed {\bf ELE-AdaptNet} yielded confident posterior probabilities,
while the other networks did not perform as intended.
This indicates that
the proposed {\bf ELE-AdaptNet} achieved reliable recognition,
irrespective of image type.
Furthermore, as the security level of the scrambled image increases, 
an adequate adaptation network
is required for a reliable recognition.

\section{Conclusion}
\label{sec:conclusion}

The block-wise scrambling algorithm was introduced herein
to hide perceptual information.
In addition,
an adaptation network was proposed to recognize such scrambled images.
Experimental comparisons implied that
the proposed block-wise scrambling and adaptation network could achieve simple perceptual information hiding
on DNN-based image analysis.


\section{Acknowledgements}
\label{sec:acknowledgements}
This work was supported by JST CREST Grant Number JPMJCR19F5.


\bibliographystyle{aaai}
\bibliography{main}
\end{document}